\documentclass[10pt,twocolumn,letterpaper]{article}

\usepackage[pagenumbers]{cvpr} % To force page numbers, e.g. for an arXiv version

\definecolor{cvprblue}{rgb}{0.21,0.49,0.74}
\usepackage[pagebackref,breaklinks,colorlinks,allcolors=cvprblue]{hyperref}

\newcommand*{\email}[1]{%
    \normalsize\href{mailto:#1}{#1}\par
    }
\title{Reliable Poisoned Sample Detection against Backdoor Attacks\\  
Enhanced by Sharpness Aware Minimization}

\author{
Mingda Zhang\textsuperscript{1}\ \ \ \ 
Mingli Zhu\textsuperscript{1} \ \ \ \ 
Zihao Zhu\textsuperscript{1} \ \ \ \ 
Baoyuan Wu\textsuperscript{1}\thanks{Corresponds to Baoyuan Wu (\email{wubaoyuan@cuhk.edu.cn}).} \\
\textsuperscript{1}School of Data Science, \\
The Chinese University of Hong Kong, Shenzhen, Guangdong, 518172, P.R. China
}

\usepackage{tabularx}
\usepackage{booktabs} % For better looking tables
\usepackage{multirow}
\usepackage{pifont}

\usepackage{amsthm}
\theoremstyle{plain} % Default style: bold title, italicized body
\newtheorem{theorem}{Theorem}[section]
\theoremstyle{definition} % Bold title, normal body

\theoremstyle{remark} % Italic title, normal body

\newtheorem{proposition}[theorem]{Proposition}

\newtheorem*{remark}{Remark}

\definecolor{darkgreen}{RGB}{0,90,0}

\begin{document}
\maketitle
\begin{abstract}

% Deep Neural Networks (DNNs) are highly susceptible to backdoor attacks, which can lead to the model performing normally on clean samples but generating specific outputs for specially crafted samples through a subset of poisoned training data. These samples often contain special triggers implanted by attackers. Filtering out these poisoned samples from the training dataset is considered a useful solution. However, these filtering methods significantly lose effectiveness against weaker backdoor attacks, such as those with lower poison rates and weaker trigger strengths, which can be manipulated by attacks easily. In this work, we first investigate the model performance of the weaker backdoor attack and find a positive correlation between the backdoor effect and detection results. Additionally, we found that sharpness aware minimization (SAM) can enhance the backdoor effect without modifying the dataset. To further understand this phenomenon, we developed an analysis of how SAM improves the backdoor effect. Consequently, we propose a novel detection framework enhanced by SAM, which can be integrated with various current detection methods to improve performance. To validate the effectiveness of this detection method, we conducted experiments on several benchmark datasets and network architectures, where our approach has demonstrated an enhancement in the performance of different detection methods under various backdoor attacks. Overall, our work provides a new perspective on enhancing the capability to detect poisoned samples.  

Backdoor attack has been considered as a serious security threat to deep neural networks (DNNs). 
Poisoned sample detection (PSD) that aims at filtering out poisoned samples from an untrustworthy training dataset has shown very promising performance for defending against data poisoning based backdoor attacks. 
However, we observe that the detection performance of many advanced methods is likely to be unstable when facing weak backdoor attacks, such as low poisoning ratio or weak trigger strength. 
To further verify this observation, we make a statistical investigation among various backdoor attacks and poisoned sample detections, showing a positive correlation between backdoor effect and detection performance. 
It inspires us to strengthen the backdoor effect to enhance detection performance. 
Since we cannot achieve that goal via directly manipulating poisoning ratio or trigger strength, we propose to train one model using the Sharpness-Aware Minimization (SAM) algorithm, rather than the vanilla training algorithm. We also provide both empirical and theoretical analysis about how SAM training strengthens the backdoor effect. 
Then, this SAM trained model can be seamlessly integrated with any off-the-shelf PSD method that extracts discriminative features from the trained model for detection, called SAM-enhanced PSD.
Extensive experiments on several benchmark datasets show the reliable detection performance of the proposed method against both weak and strong backdoor attacks, with significant improvements against various attacks ($+34.38\%$ TPR on average), over the conventional PSD methods (i.e., without SAM enhancement).
Overall, this work provides new insights about PSD and proposes a novel approach that can complement existing detection methods, which may inspire more in-depth explorations in this field.

\end{abstract}    
\section{Introduction}
\label{sec:intro}

% 我们assert 检测方法是跟强度相关的（tsne和检测的结果）

% \begin{figure}[t]
%     \centering
%     \includegraphics[width=0.8\linewidth]{figs/fig1_tsne.png}
%     \caption{Evaluating the impact of reduced poisoning ratios and intensities on backdoor attacks: a comparative analysis using t-SNE visualizations and performance metrics. The top row backdoor models with a higher poisoning ratio of 5\% and blending ratio of 0.2, while the bottom row shows results from a more subtle attack scenario with a poisoning ratio of 1\% and blending ratio of 0.1. This systematic decrease in attack hyperparameters demonstrates the resilience of ASR and the corresponding vulnerability in detection capabilities and backdoor effects quantified by TAC.}
%     \label{fig:tsne_ratio}
% \vspace{-0.7em}
% \end{figure}
\begin{figure}[t]
    \centering
    \includegraphics[width=0.8\linewidth]{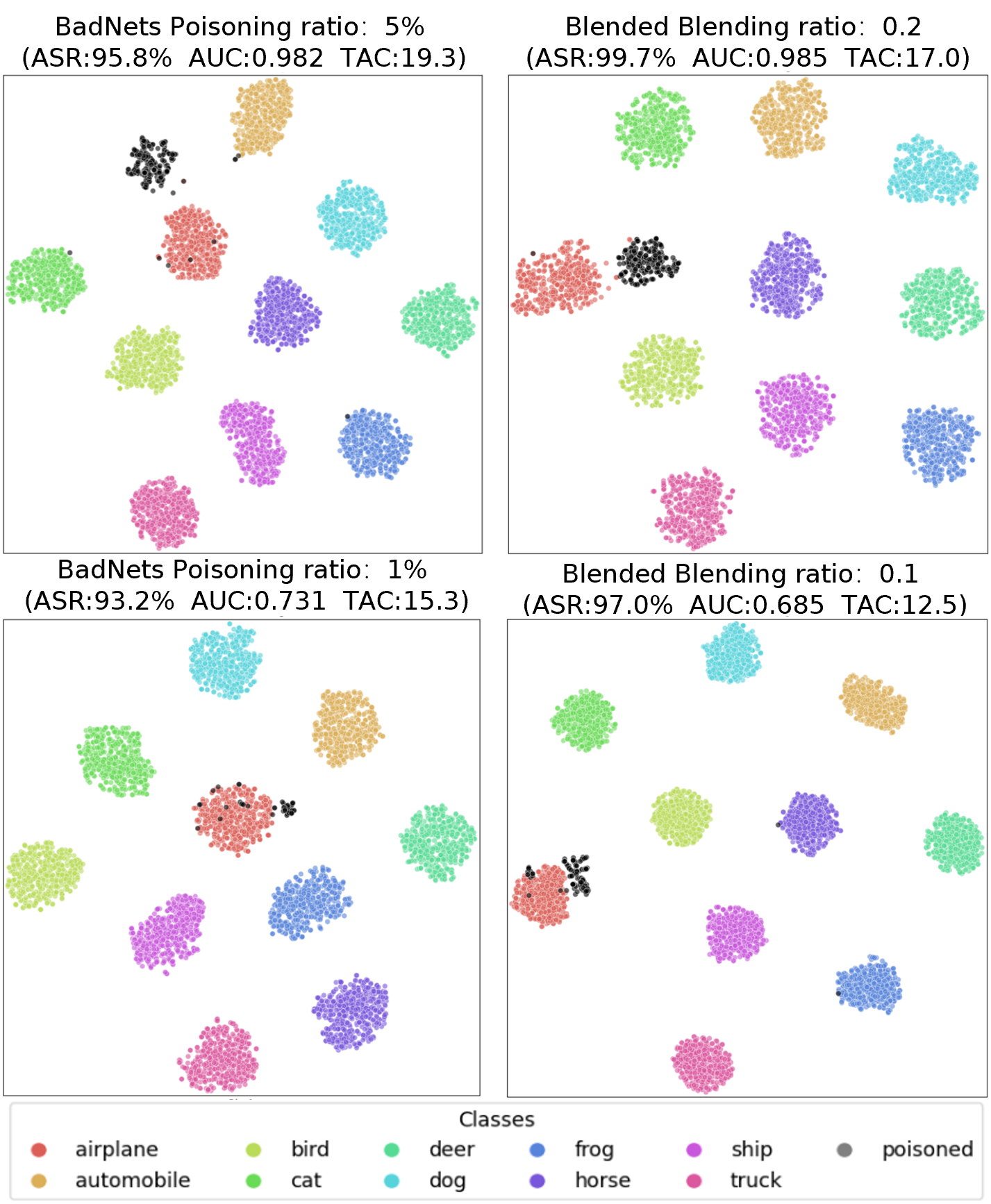}
    \vspace{-0.2cm}
    \caption{T-SNE visualizations for the impact of poisoning ratios and trigger strengths on backdoor attacks. The top row shows backdoor attacks with a higher poisoning ratio (5\%) and blending ratio (0.2), while the bottom row shows results of weak attacks with a poisoning ratio of 1\% and a blending ratio of 0.1.}
    \label{fig:tsne_ratio}
\vspace{-0.5cm}
\end{figure}
Deep Neural Networks (DNNs) are increasingly being deployed to solve complex problems across a variety of domains, including autonomous driving, medical image analysis, and social network analysis~\citep{yatbaz2023introspection,panagoulias2024evaluating,shu2024llm}. However, recent studies have shown that DNNs are vulnerable to backdoor attacks~\citep{wu2024backdoorbench}. In such attacks, the model behaves normally on clean inputs but generates targeted outputs when presented with inputs containing the trigger pre-specified by the attacker. These risks are particularly concerning when models are trained on unauthenticated third-party datasets, where malicious attackers modify a small number of samples by embedding a trigger onto them~\citep{badnet}. This poses significant security risks in practical scenarios. For example, an autonomous driving model might misinterpret a stop sign with a carefully crafted trigger as a signal to continue driving, potentially causing accidents. Therefore, it is essential to identify and eliminate these poisoned samples from the training dataset to mitigate such security risks.

Various poisoned sample detection (PSD) methods have been developed to defend against data poisoning-based backdoor attacks~\citep{wu2023defenses}.
These works typically involve training a backdoored model using the given poisoned dataset and then leveraging the performance differences between poisoned and clean samples to carry out the detection, such as sensitivity to perturbations \cite{strip}, or clustering phenomenon in feature space \cite{ac}.
While these methods have shown promising results, we observe that their performance can be unstable when faced with weak backdoor attacks, such as those with low poisoning ratios or weak trigger strength.
We highlight that \textit{the success of a PSD relies on a significant performance disparity between poisoned and clean samples, as assumed in}  \cite{spectre, strip, tact-scan}, without which poisoned samples cannot be completely distinguished from clean ones, leading to detection failure.
To verify this, we employ BadNets~\citep{badnet} and Blended~\citep{blended} attacks under different poisoning ratios and trigger strengths, and use AC~\citep{ac} and SCAn~\cite{tact-scan} for detection. 
As shown in \cref{fig:tsne_ratio}, poisoned samples are well-separated from clean samples in the feature space when the poisoning ratio or trigger strength is high, but the disparity diminishes when these factors are low, leading to detection failure (with average AUC dropping from 0.982 and 0.985 to 0.731 and 0.685, respectively).
In real-world scenarios, attackers can reduce this disparity by lowering poisoning ratios \citep{adapt} or weakening trigger strengths \citep{tact-scan,adapt,zhu2024breaking}, which hinges on the effectiveness of current PSD.

To facilitate our analysis, we first apply a concept, the \textit{backdoor effect}, to represent the impact of a trigger on clean samples within backdoor-related neurons. This effect can be measured using Trigger-Activated Change (TAC)~\citep{clp}, which quantifies the degree of backdoor relevance of neurons. As shown in~\cref{fig:tsne_ratio}, there is a noticeable decline in TAC when the poisoning ratio and trigger strength decrease. To further verify this observation, we make a statistical investigation among various backdoor attacks and PSDs and observe a positive correlation between the strength of the backdoor effect and detection performance. 
However, since defenders lack trigger information and cannot directly modify the dataset to enhance the backdoor effect, we shift our focus to optimization methods for training more robust detection models.
Inspired by the above analysis, we adopt the Sharpness-Aware Minimization (SAM) training strategy, which can amplify the backdoor effect specifically on backdoor-related neurons while simultaneously reducing its influence on other neurons in the trained model. We establish a theoretical relationship between the model trained by SAM and backdoor-related neurons. Building on these insights, we propose a novel SAM-enhanced PSD, in which the SAM-optimized model serves as a foundation for detecting poisoned samples. Our approach is both model-agnostic and data-independent, enabling seamless integration with any existing PSDs that extract discriminative features from the trained model for detection.
To prevent SAM from increasing feature variance among clean samples—which could cause unstable detection—we also use Feature-Scaling to re-weight the features, which lessens the impact of high-variance features and enhances low-variance ones, maintaining stability in detection.
Extensive experiments on ten different backdoor attacks and five detection methods across a variety of datasets and model architectures demonstrate that our approach significantly enhances the efficacy of existing detection techniques, thereby improving the robustness and reliability of PSD.
Overall, our work provides a novel approach to enhancing the detection of poisoned samples and bolstering the resilience of DNNs against backdoor attacks.

% To address this issue, we first conducted an experimental analysis to assess how the poisoning ratio and trigger strength influence the neurons associated with the backdoor, a phenomenon we refer to as the backdoor effect. This effect impacts the detection capabilities since the performance disparity between poisoned and clean samples on these neurons is larger compared to others, which can be exploited by detection methods. To enhance the backdoor effect without modifying the data, we leveraged sharpness aware minimization (SAM), which learns low-rank features~\citep{}. This approach reduces the performance of poisoned samples on other neurons while enhancing their impact on backdoor-related neurons. We also theoretically established the relationship between SAM-enhanced performance and backdoor neurons in a two-layer ReLU neural network. Based on these findings, we propose using SAM to pre-train backdoor models instead of standard training protocols, serving as the foundation for other detection methods. To mitigate the impact of SAM on the feature variance of clean samples, we introduced a scale feature module that projects the original sample features, aiding detection methods that differentiate based on features. To validate the effectiveness, we tested ten types of attacks and seven detection methods across various datasets and model architectures, improving the efficacy of existing detection methods.

In summary, our main \textbf{contributions} are: 
(1) We identify a positive correlation between the backdoor effect and detection methods, demonstrating that a weak backdoor effect can hinder the detection of poisoned samples. 
(2) We propose the integration of Sharpness-Aware Minimization (SAM) into existing PSD, which amplifies the backdoor effect in backdoored models, thereby improving the detectability of poisoned samples.
(3) Through extensive experiments, we show that our approach significantly enhances the performance of various detection methods against different attack strategies.

% We believe that this performance disparity is directly related to the backdoor effect. \textbf{Backdoor effect} refers to the extent of the trigger's impact on the performance of clean samples under the current backdoor model. When the poisoning ratio or trigger strength is high, the model focuses more on learning from poisoned samples, leading to a more pronounced performance of neurons associated with poisoned samples, thereby increasing the performance gap between poisoned and clean samples and enhancing detection efficiency. Therefore, we assert that the detection outcomes of these methods are positively correlated with the backdoor effect of poisoned samples.

% 我们的贡献

\section{Related work}
\label{sec:related_work}

\paragraph{Backdoor attack.} BadNets~\cite{badnet} introduces the concept of a backdoor attack. It randomly selects a small number of samples in the training dataset as poisoned samples, adds a small patch with a special pattern in the corner, and changes the label to the target label. In this way, the trained model will mistakenly predict the target label for any sample containing the special pattern. Other researchers propose various techniques to enhance the effectiveness of backdoor attacks. For instance, Blended~\cite{blended} employs image fusion technology, SSBA~\cite{ssba} injects specific watermarks through neural networks, and the LF~\cite{lf} applies delicate uniform modifications in the frequency domain. All these techniques aim to make backdoor attacks more covert and challenging to detect while maintaining model performance. Sleeper-agent~\cite{sleeper-agent} and Lira~\cite{lira} design more elusive poisoned samples by optimizing the poisoning method. To make poisoned samples harder to detect, the LC~\cite{lc} utilizes a different strategy, keeping the original label of the poisoned samples. This approach significantly increases the difficulty of manual detection. Concurrently, TaCT~\cite{tact-scan} and Adap-Blend~\cite{adapt} strive to enhance the similarity between poisoned and clean samples. By designing clever regularization terms, they ensure that poisoned samples are indistinguishable from clean samples even at the feature level~\cite{liang2024badclip}, effectively decreasing the chance of detection and further improving the stealthiness of the attack.

\paragraph{Backdoor defense.} Based on different stages, the backdoor defense can be divided into three stages: pre-training backdoor defense~\citep{ac,strip,adapt,beatrix}, in-training defense\citep{abl,dbr,dbd}, post-training defense~\citep{anp,rnp,ftsam,zhu2023neural,wei2023shared} and inference defense~\citep{scaleup,strip}. Pre-training defenses focus on detecting and removing suspected poisoned samples to obtain a clean dataset before training. In-training and post-training defenses aim to train models that are robust to backdoors or to eliminate backdoors from existing models. Inference-time defenses detect or repair poisoned samples during inference without altering the model. In the pre-training stage, the current main detection method is to extract information such as representation, final predictions, or gradient from the trained model, and construct discriminative metrics between poisoned and clean samples based on this information. AC~\citep{ac} distinguishes poisoned and clean samples through K-means clustering. SS~\citep{ss} and Spectre~\citep{spectre} distinguish by constructing rubust statistics through the spectrally separable hypothesis. AGPD~\citep{agpd} identifies clean samples and poisoned samples by constructing the Gradient Circular Distribution. Current detection methods often focus on identifying more effective algorithms to detect poisoned samples, but they overlook the impact of the gap between poisoned and clean samples on the detection process. We are the first to use sharpness aware minimization (SAM) training to enhance the distinction between clean and poisoned samples, thereby improving detection performance. Although FT-SAM~\citep{ftsam} also employs SAM for backdoor defense, it focuses solely on post-training backdoor defense using clean samples, which results in a reduction in the weight norm of the backdoor neurons, thus mitigating the backdoor in the model.

\paragraph{Sharpness aware minimization.} The concept of loss landscape's topology significantly influencing model generalization in deep learning has been extensively studied. A prominent hypothesis suggests that flat and wide minima in the loss landscape lead to better generalization compared to sharp minima~\citep{hochreiter1997flat}. Recent methodologies such as Sharpness-Aware Minimization (SAM) have been developed to exploit this hypothesis by concurrently minimizing the loss value and its sharpness, thereby improving model generalization~\citep{foret2020sharpness}. SAM, and its variants, employ various strategies to seek out these flat minima. For instance, ASAM~\citep{kwon2021asam} adjusts the learning method to remain invariant to parameter scaling, focusing on achieving a flat loss surface. Additionally, GSAM~\citep{zhuang2022surrogate} introduces a surrogate gap minimization alongside the perturbed loss to optimize the training process further. Beyond these, methods such as entropy-SGD and Randomized Smoothing have also been proposed to locate wider minima or enhance generalization, respectively~\citep{chaudhari2019entropy,cohen2019certified}. Theoretical and empirical analyses have provided deeper insights into the mechanisms and effectiveness of SAM. Studies have shown that SAM can lead to increased sparsity in active neurons and more compressible models, indicating a broader impact on network structure and efficiency (related studies). Moreover, SAM has been specifically effective in enhancing generalization across new architectures such as vision transformers and MLP-Mixers, suggesting its broad applicability~\citep{chen2021vision}.

\section{Method}
\label{ref:method}

\subsection{Problem setting}

% This section discusses the threat model and problem setting of backdoor attacks and detection backdoor setting.

\paragraph{Threat model.}

% We consider the scenario of data poisoning, in which the attacker only modifies the samples in the dataset, without the ability to manipulate the user's training process. Malicious attackers possess a clean training dataset $D_{cl} = \{(\mathbf{x}_i, y_i)\}_{i=1}^N \subset \mathcal{X} \times \mathcal{Y}$, where $\mathcal{X} \subset \mathbb{R}^d$ and $\mathcal{Y} = \{0,1,2,\dots, K-1\}$ represent the input space and the label set, respectively. A subset $D_{sub} \subset D_{cl}$ is used to generate the poisoned sample, The size of the subset relative to the total dataset, $p = \frac{|D_{sub}|}{|D_{cl}|}$, is referred to as the poisoning ratio. The attacker generates poisoned sample $(\tilde{\mathbf{x}}=g(x,\Delta),y_{t})$ with a trigger $\Delta$, generation function $g$ and target label $y_t$, which construct the poisoned subset $D_{poi}=\{(\tilde{\mathbf{x}}=g(\mathbf{x},\Delta),y_{t})\mid (\mathbf{x},y)\in D_{sub}\}$. The final training dataset is constructed as $D_{tr} = D_{cl} \cup D_{sub} \cup D_{poi}$, which is released by attackers to users, resulting in a backdoor in the trained models by users. This backdoor can be activated by malicious attackers, posing a security risk.
In this work, we consider data poisoning-based backdoor attack, where an attacker releases a poisoned training dataset $\mathcal{D}_{tr}$ to inject a backdoor into any model trained on it.
The attacker initially possesses a clean training dataset \( \mathcal{D}_{cl} = \{(\boldsymbol{x}_i, y_i)\}_{i=1}^N \subset \mathcal{X} \times \mathcal{Y} \), where \( \mathcal{X} \subset \mathbb{R}^d \) and \( \mathcal{Y} = \{0,1,2,\dots, K-1\} \) represent the input space and label set, respectively. A subset \( \mathcal{D}_{sub} \subset \mathcal{D}_{cl} \) is selected to generate poisoned samples, and the proportion of this subset relative to the total dataset, \( p = \frac{|\mathcal{D}_{sub}|}{|\mathcal{D}_{cl}|} \), is defined as the \textit{poisoning ratio}. Using a pre-defined trigger \( \Delta \), a generation function \( g \), and a target label \( y_t \), the attacker constructs each poisoned sample \( (\tilde{\boldsymbol{x}} = g(\boldsymbol{x}, \Delta), y_t) \), forming the poisoned subset \( \mathcal{D}_{poi} = \{(\tilde{\boldsymbol{x}}, y_t) \mid (\boldsymbol{x}, y) \in \mathcal{D}_{sub}\} \). The final poisoning dataset is then created as \( \mathcal{D}_{tr} = \mathcal{D}_{cl} \setminus \mathcal{D}_{sub} \cup \mathcal{D}_{poi} \) and released to users, embedding a backdoor into any trained models.

\paragraph{Defender's goal.}

% Defenders, lacking information about the attacker's actions such as the poisoning rate $p$, the trigger $\Delta$, and the generation function $g$, detect the poisoned sample in the released training dataset. Additionally, defenders can access a small clean subset $D_{sur} \subset \mathcal{X} \times \mathcal{Y}$, which is drawn from the same distribution as the original training dataset. This subset serves as a reference for detecting poisoned samples.

Defenders aim to detect the poisoned samples $\mathcal{D}_{poi}$ within the released training dataset. They lack attack details such as the poisoning ratio $p$, the trigger $\Delta$, and the generation function $g$. We assume defenders can access a few clean samples, which is drawn from the same distribution as $\mathcal{D}_{cl}$. This subset serves as a reference for detecting poisoned samples.

\subsection{Correlation between backdoor effect and detection performance}
\label{sec:pre}
\paragraph{Definition of TAC.}

Trigger Activation Change (TAC) metric, introduced by \citet{clp}, primarily measures the differences in neuronal activation values of DNN by comparing poisoned samples to the corresponding clean samples. We denote a deep neural network (DNN) as $f_{\boldsymbol{\theta}} = f^{(L)} \circ f^{(L-1)} \cdots f^{(2)} \circ f^{(1)}$. For some clean samples $\boldsymbol{x}$ and their corresponding poisoned versions $\tilde{\boldsymbol{x}}$, TAC is computed through the following equation:
\begin{align}
    T A C_k^{(l)}(\mathcal{D})=\frac{1}{|\mathcal{D}|} \sum_{\boldsymbol{x} \in \mathcal{D}}\left\|f_k^{(l)}(\boldsymbol{x})-f_k^{(l)}(\tilde{\boldsymbol{x}})\right\|_2,
\end{align}
where $k$ represents the $k_{th}$ neuron in layer $l$, and $\mathcal{D}$ is the set of clean input samples.

% In this section, we introduce TAC (Trigger Activation Clustering) as a numerical indicator to measure the effect of backdoors.
% Comparison between Top-k TAC and AUC across different backdoor attacks and detections. 这些后门攻击是在CIFAR-10上训练的，使用的是ResNet18，包含三种poisoning ratio$\{0.5\%,1\%,5\%\}$
\begin{figure}[t]
    \centering
    \includegraphics[width=1.0\linewidth]{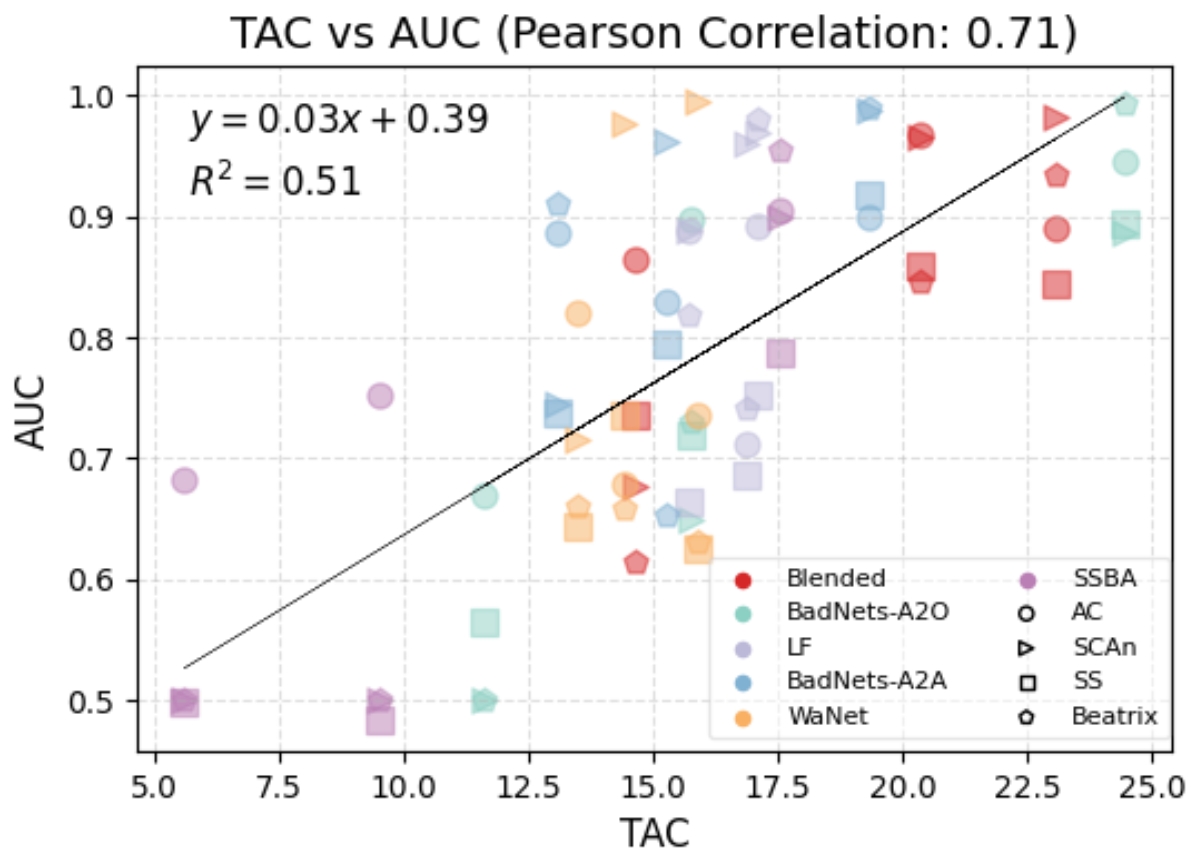}
    \vspace{-0.4cm}
    \caption{Comparison between Top-K TAC and AUC across various backdoor attacks and detections. These backdoor attacks are trained on the CIFAR-10 dataset and ResNet18 where $K=2$, including three poisoning ratios: $\{0.5\%, 1\%, 5\%\}$. Distinct shapes and colors denote various detection and attack methods, respectively.}
    \label{fig:tac_auc}
\vspace{-0.6cm}
\end{figure}

\paragraph{Backdoor effect and detection performance.}

% Based on the definition of the TAC, we can see that the magnitude of this difference indicates the trigger's effect on a specific neuron. A higher TAC indicates that the neuron is more sensitive to the trigger, leading to a more pronounced response, which be considered as backdoor neurons. The performance of all these backdoor neurons can be regarded as the \textbf{backdoor effect} of the model, which can be calculated by taking the mean of the Top-K TAC values. We hypothesize that the stronger the backdoor effect, the greater the disparities between poisoned and clean samples, which can improve the detection results.

% To validate our hypothesis, we conduct a series of experiments focusing on various types of attacks. We control the backdoor effect by adjusting the poisoning rate or the strength of the trigger and employ the Top-K TAC to gauge the backdoor model response and the True Positive Rate (TPR) of different detection methods to measure accuracy in identifying poisoned samples with similar False Positive Rate (FPR). As illustrated in \cref{fig:tac_ratio}, the results indicate that as the poisoning rate and trigger strength increase, the mean value of the Top-K TAC also rises, suggesting that the impact of the trigger on the model increases, and detection methods become more effective. It not only validates our hypothesis but also provides directions for improving detection methods, i.e., how to increase the backdoor effect without prior knowledge of the backdoor attack only by altering the training process. 

According to the definition of the TAC, the magnitude of this difference indicates the trigger's effect on a specific neuron. A higher TAC indicates that the neuron is more sensitive to the trigger, leading to a stronger response and thus can be identified as a backdoor neuron. 
The collective performance of these backdoor neurons constitutes the model’s backdoor effect, which we quantify by taking the mean of the Top-K TAC values. \textit{We hypothesize that a stronger backdoor effect leads to a greater disparity between poisoned and clean samples, thereby improving detection performance.}

To test our hypothesis, we adjust the poisoning ratio to control backdoor effects and evaluate different detection methods under various attack types on CIFAR-10 and ResNet18. We use the Top-K TAC to assess the model's backdoor effect and evaluate detection performance by measuring the Area Under Curve (AUC), a commonly used statistical measure to indicate the predictive ability of detection. As shown in \cref{fig:tac_auc}, the detection performance declines as the TAC decreases, and the Pearson correlation coefficient between these two metrics is 0.71, which means there is a positive correlation between the backdoor effect and detection performance. Additionally, the regression analysis resulted in an R-squared value of 0.51, indicating that TAC, as an independent variable, has explanatory power over the variation in AUC. Our findings not only validate the hypothesis but also suggest a promising direction for enhancing detection methods—namely, increasing the backdoor effect by adjusting the training process, even without prior knowledge of the specific backdoor attack.

\subsection{Backdoor attack with SAM}

% \paragraph{Motivation} \citet{adapt} discuss that backdoor attacks constitute a form of shortcut learning. By weakening the backdoor trigger to make the model learn more semantic information, the difference between backdoor and clean samples can be reduced. Given that backdoor attacks leverage shortcut learning, our analysis confirms that detection results are indeed more closely associated with the activation of backdoor neurons. We can suppress the learning of redundant features during backdoor learning, thereby concentrating the learning process more on backdoor features to enhance the backdoor effect. \citet{sam_lowrank} mention that Sharpness Aware Minimization (SAM) achieves a lower rank of features by suppressing certain features. Since SAM enables the model to focus more on backdoor characteristics, the model can learn a stronger backdoor signal.

\paragraph{Effect of SAM on backdoor effect.} 

% $\mathcal(L)(\boldsymbol{\theta})=\frac{1}{|D_{cl}-D_{sub}|}\sum_{(\mathbf{x},y)\in (D_{cl}-D_{sub})}\ell_{ce}(f_{\boldsymbol{\theta}}(\mathbf{x});y) + \frac{1}{|D_{poi}|}\sum_{(\tilde{\mathbf{x}},y_t)\in (D_{poi})}\ell_{ce}(f_{\boldsymbol{\theta}}(\tilde{\mathbf{x}};y_t)$

% 首先我们先介绍后门攻击时，使用的损失函数$\mathcal{L}(\boldsymbol{\theta})$去保证模型本身能力的情况下，学习到后门样本的特征，which is defined as follows:

% 让我们先介绍poisoned sample的损失函数，which is defined as follows:

% 为了增强后门effect，对于后门损失，我们使用SAM优化函数，which can be represented as follows:

\begin{figure}[t]
    \centering
    \includegraphics[width=1.0\linewidth]{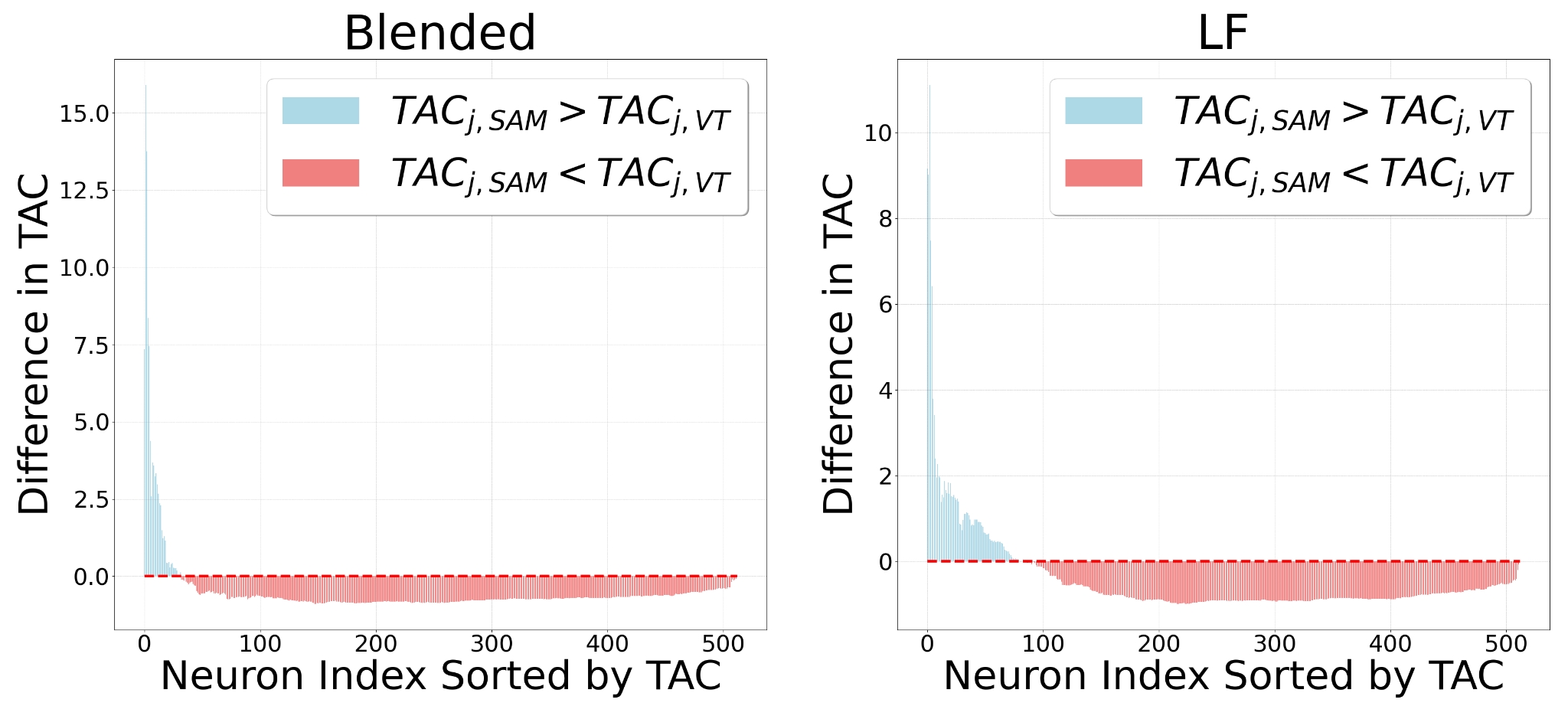}
    \vspace{-0.2cm}
    \caption{The differences in all TACs between the model trained with SAM and the model trained with Vanilla Training. These neurons are indexed in descending order based on the TAC in their respective models, which means that a smaller index indicates that this pair of neurons has a higher TAC in the corresponding model.}
    \label{fig:diff_tac}
\vspace{-0.5cm}
\end{figure}

To enhance the backdoor effect, we utilize the Sharpness-Aware Minimization (SAM) optimization for training the backdoored model. It is formulated as follows:
\begin{align}
\label{eq:sam}
\min_{\boldsymbol{\theta}}\max_{\boldsymbol{\epsilon}\in\{\|\boldsymbol{\epsilon}\|_{2} \leq \rho\}}\mathcal{L}(\boldsymbol{\theta}+\boldsymbol{\epsilon}),
\end{align}
where $\mathcal{L}(\boldsymbol{\theta})=\frac{1}{|\mathcal{D}_{tr}|}\sum_{(\boldsymbol{x},y)\in \mathcal{D}_{tr}}\ell(f_{\boldsymbol{\theta}}(\boldsymbol{x};y)$ is the cross-entropy loss, and $\rho > 0$ is a hyperparameter that controls the budget for weight perturbations. To illustrate the effect of SAM algorithm, we conduct experiments to compare TAC values before and after applying SAM under various attacks.
As illustrated in \cref{fig:diff_tac}, it is evident that for neurons with high TAC values (\ie, backdoor neurons), their TAC values are significantly increased. Conversely, for neurons with low TAC values  (\ie, neurons unrelated to the backdoor), SAM can even decrease their TAC values. \textit{This suggests that SAM amplifies its impact on backdoor neurons by forcing the model to learn and activate the most prominent backdoor features.}

\paragraph{Theoretical Analysis of SAM's Effect.}

In this section, we theoretically analyze the impact of SAM on backdoor effect.
To simplify our analysis, we focus on a binary classification task using a two-layer ReLU neural network, defined as $f(\boldsymbol{\theta}) = \boldsymbol{a} * \sigma(\boldsymbol{Wx})$, where $\sigma$ denotes the activation function. The cross-entropy loss $\ell(\boldsymbol{\theta})$ with a sample $(\boldsymbol{x}, y)$ is utilized to train the model. Without loss of generality, we assume that the target label for a poisoned sample $\tilde{\boldsymbol{x}}$ is 0. We present our theoretical argument in the following proposition.

\begin{proposition}
For any activated neuron in a two-layer ReLU network trained with cross-entropy loss, each update via SAM increases its pre-activation values $\{\left\langle \boldsymbol{w}_j, \tilde{\boldsymbol{x}}\right\rangle\}_{j=1}^m$ with respect to SGD according to a poisoned sample given the condition $a_{j} \sigma^{\prime}\left(\left\langle \boldsymbol{w}_j, \tilde{\boldsymbol{x}}\right\rangle\right) < -\frac{\sigma\left(\left\langle \boldsymbol{w}_j, \tilde{\boldsymbol{x}}\right\rangle\right)}{(1-\ell'(\boldsymbol{\theta}))\|\nabla f(\boldsymbol{\theta})\|^2_2}$.
\end{proposition}

\begin{remark}
For neurons that meet this condition, we observe that $a_{j} < 0$ and it is activated by the poisoned sample $\tilde{\boldsymbol{x}}$ in the process. Given our assumption that $y_t = 0$, this implies a higher probability that the neuron’s output for the poisoned sample is not zero the more negative $a_{j}$ becomes. Meanwhile, neurons satisfying these conditions have significantly negative weights $a_{j}$. Hence, neurons that meet the condition are highly related to the poisoned sample, which means they can be identified as backdoor neurons. In summary, compared to Vanilla training (SGD), SAM increases the activation of poisoned samples on the backdoor neuron, thereby enhancing the backdoor effect.
\end{remark}

\subsection{SAM-enhanced PSD}

\begin{figure}[t]
    \centering
    \includegraphics[width=1.0\linewidth]{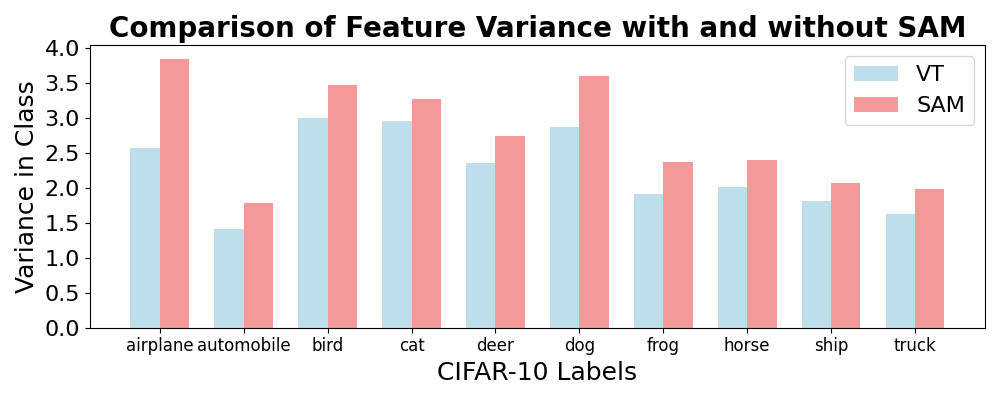}
    \vspace{-0.5cm}
    \caption{Comparison of the intra-class feature variance between the model trained with SAM and the model trained with Vanilla Training.}
    \label{fig:var}
\vspace{-0.6cm}
\end{figure}

Inspired by the previous two insights—that the backdoor effect is positively correlated with detection efficacy, and that SAM enhances the backdoor effect—we propose SAM-enhanced Poison Sample Detection (PSD), which consists of three stages:
\begin{itemize}
    \item \textbf{Stage-1: Training Backdoor Model.} Train a backdoor model $f_{\boldsymbol{\theta_{\text{SAM}}}}$ using SAM according to \cref{eq:sam}.
    \item \textbf{Stage-2: Computing Features.} Extract features from the current model $f_{\boldsymbol{\theta_{\text{SAM}}}}$ by computing $\boldsymbol{g} = \phi_{\boldsymbol{\theta_{\text{SAM}}}}(\boldsymbol{x})$ where $\phi_{\boldsymbol{\theta_{\text{SAM}}}}$ is the feature extractor of the model $f_{\boldsymbol{\theta_{\text{SAM}}}}$. Scale the feature through the feature-scaling, which contains a projection matrix $\boldsymbol{P}$ estimated from the training dataset using Principal Component Analysis (PCA) and a covariance matrix $\boldsymbol{\Sigma}$ estimated from reference clean samples and potential clean samples dynamically collected from the poisoned dataset, to address the increased intra-class variance of clean samples after SAM optimization, as shown in \cref{fig:var}. and get the scaled feature $\boldsymbol{g}^{s} = \boldsymbol{\Sigma}^{-1/2} \boldsymbol{P} \boldsymbol{g}$.
    \item \textbf{Stage-3: Integrading with off-the-shell PSD.} Call any off-the-shelf PSD method, such as Activation Clustering (AC), and use the scaled features $\boldsymbol{g}^{s}$ to detect poisoned samples.
\end{itemize}
The specific algorithmic process of SAM-enhanced PSD is detailed in the \textbf{supplementary material}.

\section{Experiment}
\label{ref:experiment}

\subsection{Experimental setup}

\paragraph{Attack settings.}
In this study, we evaluate the efficacy of backdoor attacks within an experimental framework. Specifically, we include ten backdoor attack methods: BadNets~\citep{badnet}, both in its class-specific (BadNets-A2O) and universal forms (BadNets-A2A), Blended attack~\citep{blended}, Label-consistent attack (LC)~\citep{lc}, Low-frequency attack (LF)~\citep{lf}, Sample-specific backdoor attack (SSBA)~\citep{ssba}, Targeted contamination attack (TaCT)~\citep{tact-scan}, Adaptive-Blend attack (Adap-Blend)~\citep{adapt}, Trojan attack (TrojanNN)~\citep{trojannn}, and Warping-based attack (WaNet)~\citep{wanet}. Each attack is configured according to the default settings provided by BackdoorBench~\citep{backdoorbench}. The experimental evaluation is conducted on three benchmark datasets: CIFAR-10~\citep{cifar10}, Tiny ImageNet~\citep{imagenet}, and GTSRB~\citep{gtsrb}, and implemented on three neural network architectures, namely ResNet18~\cite{resnet}, VGG19-BN~\cite{vgg} and DenseNet-161~\citep{densenet}. Due to space constraints, the results for Tiny, VGG19-BN and DenseNet-161 are presented to \textbf{supplementary material}. For our experiments, we set the poisoning ratio uniformly at 5\% across all attack types, and the target label of BadNets-A2A is reassigned to $y_t = (y + 1) \mod K$ for each class y, where K representing the total number of classes. However, for the LC attacks, which only poison clean samples with target label, can only be implemented for CIFAR-10. We consider \textbf{weak backdoor attacks} as those whose poisoning ratio is low e.g., 1\%, 0.5\%, or 0.1\%—or those with weak strength, such as Adap-Blend.

\begin{table*}[h]
\centering
\captionsetup{font=small}
\caption{Comparison of TPR (\%) and FPR (\%) between base PSD and SAM-enhanced PSD (+SAM) on CIFAR-10 and ResNet18. Top 1 are \textbf{bold}. When comparing SAM-enhanced PSD to base PSD, we highlight performance improvements in \textcolor{darkgreen}{green} and declines in \textcolor{red}{red}.}
\vspace{-0.2cm}
\label{tab:cifar10}
\renewcommand\arraystretch{1}
\resizebox{\textwidth}{!}{%
\begin{tabular}{l|ccc|ccc|ccc|ccc|ccc}
\toprule
Detection $\rightarrow $ & \multicolumn3{c|}{Spectre / +SAM} & \multicolumn3{c|}{SCAn / +SAM} & \multicolumn3{c|}{SS / +SAM} & \multicolumn3{c|}{AC / +SAM} & \multicolumn3{c}{Beatrix / +SAM} \\
 Method $\downarrow$  & TPR $\uparrow$ & FPR $\downarrow$ & F1 $\uparrow$ & TPR $\uparrow$ & FPR $\downarrow$ & F1 $\uparrow$ & TPR $\uparrow$ & FPR $\downarrow$ & F1 $\uparrow$ & TPR $\uparrow$ & FPR $\downarrow$ & F1 $\uparrow$ & TPR $\uparrow$ & FPR $\downarrow$ & F1 $\uparrow$ \\
\midrule
BadNets & 51.1/\textbf{88.4} & 4.9/\textbf{2.9} & 42.0/\textbf{72.6} & \textbf{96.0}/95.2 & 0.0/\textbf{0.0} & \textbf{98.0}/97.6 & 70.8/\textbf{92.3} & 2.4/\textbf{1.2} & 65.6/\textbf{85.5} & \textbf{96.8}/95.4 & \textbf{0.1}/13.3 & \textbf{97.1}/42.5 & 56.6/\textbf{98.8} & 5.0/\textbf{0.5} & 44.9/\textbf{94.5} \\
Blended & 29.9/\textbf{59.7} & 6.0/\textbf{4.4} & 24.6/\textbf{49.0} & \textbf{99.2}/98.7 & 0.0/\textbf{0.0} & \textbf{99.6}/99.3 & 32.9/\textbf{94.6} & 4.4/\textbf{1.1} & 30.5/\textbf{87.6} & 2.3/\textbf{98.8} & \textbf{7.7}/11.9 & 1.9/\textbf{46.4} & 5.0/\textbf{99.8} & 5.0/\textbf{1.5} & 5.0/\textbf{87.6} \\
SSBA & 36.6/\textbf{72.8} & 5.6/\textbf{3.7} & 30.1/\textbf{59.8} & 93.9/\textbf{96.5} & 0.0/\textbf{0.0} & 96.9/\textbf{98.2} & 80.4/\textbf{89.9} & 1.9/\textbf{1.4} & 74.5/\textbf{83.3} & \textbf{99.3}/96.5 & \textbf{3.3}/16.2 & \textbf{76.1}/38.3 & 16.8/\textbf{98.9} & 5.0/\textbf{0.4} & 15.8/\textbf{95.5} \\
LF & 32.0/\textbf{54.1} & 5.9/\textbf{4.7} & 26.3/\textbf{44.5} & 94.1/\textbf{96.1} & 0.0/\textbf{0.0} & 97.0/\textbf{98.0} & 68.2/\textbf{86.5} & 2.5/\textbf{1.5} & 63.2/\textbf{80.2} & 95.6/\textbf{96.1} & \textbf{10.4}/10.5 & 48.7/\textbf{48.7} & 2.4/\textbf{98.8} & 5.0/\textbf{0.8} & 2.4/\textbf{92.3} \\
Adap-Blend & 24.1/\textbf{65.9} & 5.6/\textbf{3.7} & 20.9/\textbf{56.0} & 92.5/\textbf{97.3} & 10.5/\textbf{10.5} & 47.3/\textbf{49.1} & 20.2/\textbf{91.0} & 4.5/\textbf{1.2} & 19.6/\textbf{85.4} & 1.5/\textbf{97.1} & \textbf{7.1}/7.6 & 1.2/\textbf{57.0} & 6.2/\textbf{99.9} & \textbf{5.0}/8.2 & 6.2/\textbf{56.2} \\
LC & 17.0/\textbf{41.7} & 4.3/\textbf{3.0} & 17.1/\textbf{41.9} & \textbf{100.0}/99.9 & 0.0/\textbf{0.0} & \textbf{100.0}/99.9 & 40.5/\textbf{47.8} & 2.1/\textbf{1.7} & 45.0/\textbf{53.2} & 0.0/\textbf{100.0} & 0.0/\textbf{0.0} & 0.0/\textbf{100.0} & 2.2/\textbf{99.9} & 5.0/\textbf{3.2} & 2.2/\textbf{76.6} \\
TaCT & 36.1/\textbf{78.6} & 7.0/\textbf{2.3} & 26.9/\textbf{70.9} & 100.0/\textbf{100.0} & 0.0/\textbf{0.0} & 100.0/\textbf{100.0} & 42.3/\textbf{46.4} & 4.2/\textbf{3.7} & 38.1/\textbf{42.7} & 100.0/\textbf{100.0} & 0.1/\textbf{0.0} & 99.5/\textbf{100.0} & 13.4/\textbf{100.0} & 5.0/\textbf{2.6} & 12.9/\textbf{80.2} \\
TrojanNN & 30.2/\textbf{62.4} & 6.0/\textbf{4.3} & 24.8/\textbf{51.3} & 100.0/\textbf{100.0} & 0.0/\textbf{0.0} & 100.0/\textbf{100.0} & 63.4/\textbf{97.2} & 2.8/\textbf{1.0} & 58.7/\textbf{90.1} & 99.9/\textbf{100.0} & \textbf{3.2}/12.0 & \textbf{76.7}/46.7 & 4.6/\textbf{100.0} & 5.0/\textbf{3.5} & 4.6/\textbf{75.3} \\
WaNet & 66.4/\textbf{97.7} & 4.1/\textbf{2.6} & 54.3/\textbf{79.2} & 66.3/\textbf{90.1} & 0.0/\textbf{0.0} & 79.7/\textbf{94.8} & 71.1/\textbf{86.0} & 1.5/\textbf{0.7} & 71.4/\textbf{85.9} & 85.1/\textbf{90.1} & 0.0/\textbf{0.0} & 91.9/\textbf{94.8} & 1.2/\textbf{95.5} & 5.0/\textbf{5.0} & 1.2/\textbf{65.7} \\
BadNets-A2A & 99.5/\textbf{99.6} & 10.6/\textbf{10.5} & 49.7/\textbf{49.8} & 0.0/\textbf{0.0} & 0.0/\textbf{0.0} & 0.0/\textbf{0.0} & 99.4/\textbf{99.4} & 10.6/\textbf{10.6} & 49.7/\textbf{49.7} & \textbf{97.8}/96.1 & 0.0/\textbf{0.0} & \textbf{98.9}/98.0 & 27.3/\textbf{99.2} & 5.0/\textbf{1.0} & 24.6/\textbf{91.3} \\
\midrule
Average& \textcolor{darkgreen}{+29.8} & \textcolor{darkgreen}{--1.8} & \textcolor{darkgreen}{+25.8}& \textcolor{darkgreen}{+3.2} & \textcolor{darkgreen}{-0.0} & \textcolor{darkgreen}{+1.9}& \textcolor{darkgreen}{+24.2} & \textcolor{darkgreen}{--1.3} & \textcolor{darkgreen}{+22.8}& \textcolor{darkgreen}{+29.2} & \textcolor{red}{+4.0} & \textcolor{darkgreen}{+8.0}& \textcolor{darkgreen}{+85.5} & \textcolor{darkgreen}{--2.3} & \textcolor{darkgreen}{+69.5}\\
\bottomrule
\end{tabular}
}
\vspace{-0.2cm}
\end{table*}
\begin{table*}[h]
\centering
\captionsetup{font=small}
\caption{Comparison of TPR (\%) and FPR (\%) for base poisoned sample detection with SAM-enhanced (SAM) on Tiny and ResNet18. Top 1 are \textbf{bold}. When comparing SAM-enhanced PSD to base PSD, we highlight performance improvements in \textcolor{darkgreen}{green} and declines in \textcolor{red}{red}.}
\label{tab:gtsrb}
\renewcommand\arraystretch{1}
\vspace{-0.2cm}
\resizebox{\textwidth}{!}{%
\begin{tabular}{l|ccc|ccc|ccc|ccc|ccc}
\toprule
Detection $\rightarrow $ & \multicolumn3{c|}{Spectre / +SAM} & \multicolumn3{c|}{SCAn / +SAM} & \multicolumn3{c|}{SS / +SAM} & \multicolumn3{c|}{AC / +SAM} & \multicolumn3{c}{Beatrix / +SAM} \\
 Method $\downarrow$  & TPR $\uparrow$ & FPR $\downarrow$ & F1 $\uparrow$ & TPR $\uparrow$ & FPR $\downarrow$ & F1 $\uparrow$ & TPR $\uparrow$ & FPR $\downarrow$ & F1 $\uparrow$ & TPR $\uparrow$ & FPR $\downarrow$ & F1 $\uparrow$ & TPR $\uparrow$ & FPR $\downarrow$ & F1 $\uparrow$ \\
 \midrule
BadNets & - & - & - & \textbf{97.8}/91.2 & 0.0/\textbf{0.0} & \textbf{98.9}/95.4 & - & - & - & \textbf{96.4}/91.0 & \textbf{0.2}/0.3 & \textbf{96.3}/92.7 & 25.8/\textbf{99.8} & 5.1/\textbf{5.1} & 23.2/\textbf{67.5} \\
Blended & - & - & - & 88.9/\textbf{99.6} & 0.0/\textbf{0.0} & 94.1/\textbf{99.8} & - & - & - & 0.0/\textbf{99.7} & 0.2/\textbf{0.2} & 0.0/\textbf{98.4} & 46.9/\textbf{100.0} & 5.1/\textbf{5.1} & 38.6/\textbf{67.6} \\
SSBA & - & - & - & \textbf{100.0}/97.3 & 0.0/\textbf{0.0} & \textbf{100.0}/98.7 & - & - & - & 91.0/\textbf{97.4} & \textbf{0.1}/0.3 & 94.0/\textbf{96.0} & 35.7/\textbf{100.0} & 5.1/\textbf{5.0} & 30.8/\textbf{67.6} \\
LF & - & - & - & \textbf{91.2}/85.8 & 0.0/\textbf{0.0} & \textbf{95.4}/92.3 & - & - & - & 0.0/\textbf{87.9} & \textbf{0.2}/1.3 & 0.0/\textbf{82.9} & 32.5/\textbf{99.5} & 5.1/\textbf{5.1} & 28.4/\textbf{67.4} \\
Adap-Blend & - & - & - & \textbf{99.8}/97.6 & 13.5/\textbf{0.0} & 43.7/\textbf{98.8} & - & - & - & 88.3/\textbf{96.8} & 0.3/\textbf{0.3} & 90.9/\textbf{95.3} & 97.9/\textbf{99.9} & 4.5/\textbf{4.5} & 69.0/\textbf{69.9} \\
TrojanNN & - & - & - & 99.9/\textbf{99.9} & 0.0/\textbf{0.0} & 100.0/\textbf{100.0} & - & - & - & 98.3/\textbf{99.9} & 0.4/\textbf{0.3} & 95.4/\textbf{97.2} & 36.8/\textbf{100.0} & 5.1/\textbf{5.0} & 31.6/\textbf{67.6} \\
WaNet & - & - & - & 0.0/\textbf{71.1} & 0.0/\textbf{0.0} & 0.0/\textbf{83.1} & - & - & - & 0.0/\textbf{72.6} & \textbf{0.1}/0.7 & 0.0/\textbf{78.0} & 6.3/\textbf{86.5} & 5.1/\textbf{5.1} & 6.2/\textbf{61.2} \\
BadNets-A2A & 89.7/\textbf{93.5} & 30.6/\textbf{30.4} & 23.3/\textbf{24.3} & 0.0/\textbf{0.0} & 0.0/\textbf{0.0} & 0.0/\textbf{0.0} & 92.3/\textbf{93.9} & 11.0/\textbf{10.9} & 46.1/\textbf{46.9} & 92.1/\textbf{94.2} & 0.0/\textbf{0.0} & 95.9/\textbf{97.0} & 73.7/\textbf{100.0} & \textbf{5.0}/5.1 & 54.7/\textbf{67.6} \\
\midrule
Average& \textcolor{darkgreen}{+3.8} & \textcolor{red}{+0.2} & \textcolor{darkgreen}{+1.0}& \textcolor{darkgreen}{+8.1} & \textcolor{red}{+1.7} & \textcolor{darkgreen}{+17.0}& \textcolor{darkgreen}{+1.6} & \textcolor{red}{+0.1} & \textcolor{darkgreen}{+0.8}& \textcolor{darkgreen}{+34.1} & \textcolor{darkgreen}{--0.2} & \textcolor{darkgreen}{+33.1}& \textcolor{darkgreen}{+53.8} & \textcolor{darkgreen}{-0.0} & \textcolor{darkgreen}{+31.7}\\
\bottomrule
\end{tabular}
}
\vspace{-0.5cm}
\end{table*}
\paragraph{Detection settings.} 
In this study, we systematically evaluate the effectiveness of our proposed SAM-enhanced PSD, combined with a wide range of backdoor detection methods including Activation Clustering (AC)~\citep{ac}, Beatrix~\citep{beatrix}, SCAn~\citep{tact-scan} Spectral Signature (SS)~\citep{ss} and Spectre~\cite{spectre}; Additionally, it is presumed that a small, clean dataset can be utilized to aid the detection process, a common practice in recent studies~\citep{beatrix,strip,cd,tact-scan}. For a balanced evaluation, each class in this auxiliary clean dataset contains 250 samples, which are carefully selected from the test dataset. The SAM-enhanced PSD utilizes sharpness aware minimization with perturbation budget $\rho=0.1$.

\paragraph{Evaluation metrics.} In this work, the metrics used by the defender are True Positive Rate (TPR), False Positive Rate (FPR) and F1 score. In the tables presenting our results, the top performer is highlighted in \textbf{bold}.
% F1 score is a harmonic mean of Precision and Recall. 

\begin{figure*}[htbp]
    \centering
    \includegraphics[width=1.0\linewidth]{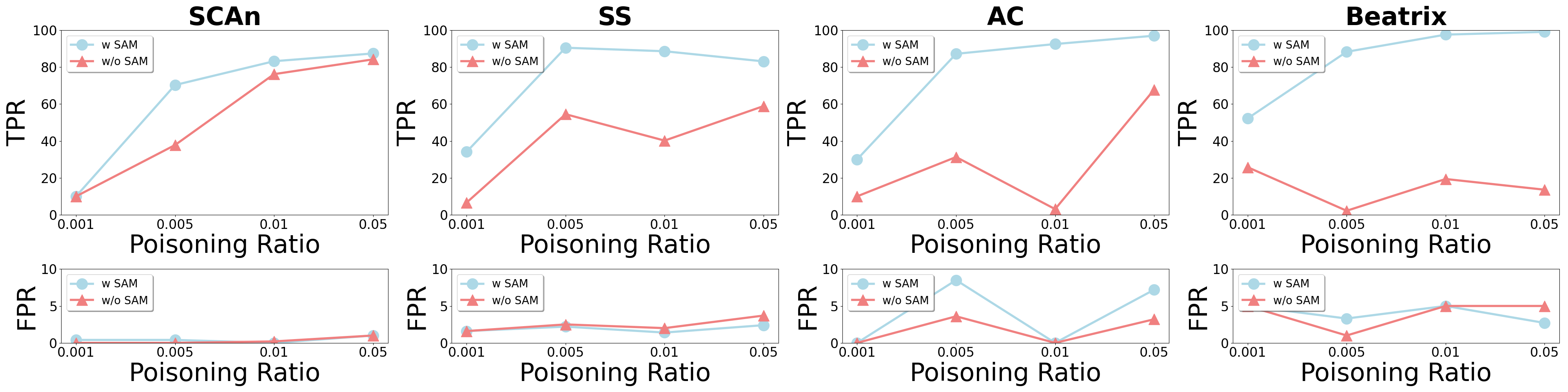}
    \vspace{-0.7cm}
    \caption{Detection performance of base PSD with SAM-enhanced PSD (SAM) under different poisoning ratios on CIFAR10 and ResNet18.}
    \label{fig:pratio}
\vspace{-0.6cm}
\end{figure*}

\subsection{Main results}
\paragraph{Effectiveness of SAM-enhanced PSD.}

To validate the effectiveness of the SAM-enhanced PSD, we demonstrated the effects of different PSDs as well as these combined with SAM-enhanced PSD on CIFAR-10 and GTSRB datasets, as shown in \cref{tab:cifar10} and \cref{tab:gtsrb}, respectively. \ding{182} The SAM-enhanced PSD generally enhances various base off-the-shell PSDs, as indicated in \cref{tab:cifar10} and \cref{tab:gtsrb}. For CIFAR-10, we improved the True Positive Rate (TPR) by over 25\% for four detection methods. Similarly, on GTSRB, we enhanced the detection performance of most methods, with increases in TPR exceeding 30\% for two methods. \ding{183} For methods like Spectre, SS and Beatrix, which are based on anomaly detection, SAM-enhanced PSD increases the prominence of poisoned samples at backdoor neurons relative to clean samples, making these samples more anomalous and thus enhancing detection. In the context of GTSRB, due to poisoned samples being closely matched to target clean samples, it breaks the assumptions of methods Spectre and SS, therefore they are not evaluated. \ding{184} For the SCAn method, our approach shows significant improvements, especially under attacks where SCAn typically underperforms, such as WaNet. Since SCAn requires identification of the target label, it cannot detect BadNets-A2A attacks. \ding{185} For the AC, since SAM-enhanced PSD  increases the activation of poisoned samples in backdoor-related neurons, it causes poisoned samples to deviate more from clean samples and cluster more tightly, thereby enhancing detection effectiveness.

\vspace{-0.2cm}
\paragraph{Performance under different poisoning ratios.}

To evaluate the impact of the poisoning ratio on the SAM-enhanced PSD, especially in the case of what we consider to be \textbf{weak backdoor attacks}, we present the average detection performance of four detection methods under all attacks on CIFAR-10 and ResNet18, as shown in \cref{fig:pratio}. The selected range for the poisoning ratio is \{0.1\%, 0.5\%, 1\%, 5\%\}. \ding{182} SAM-enhanced PSD enhances detection performance across different poisoning ratios as illustrated in \cref{fig:pratio}. Notably, when the poisoning ratio is low such as $0.5\%$ and $1\%$, SAM-enhanced PSD significantly improves the performance of PSDs. \ding{183} When the poisoning ratio is 0.1\%, even though SAM-enhanced PSD improves performance, its average True Positive Rate (TPR) does not exceed 60\%. The average attack success rate is only 31.8\%, meaning a complete backdoor cannot form, which makes poisoned samples harder to detect. Detailed results are provided in the \textbf{supplementary materials}.

\subsection{Ablation study}

\paragraph{Effect of each component in SAM-enhanced PSD.}

As shown in \cref{tab:abla}, we evaluate the impact of sharpness-aware minimization (SAM) and feature-scaling (FS) on detection backdoor attacks, specifically BadNets and Blended, using SS and Beatrix detection methods. Integrating the SAM significantly enhances detection effectiveness by increasing the feature gap between clean and backdoor samples. The addition of the FS further improves detection by compressing features and reducing noise in clean sample features, leading to clearer distinctions.  When SAM and FS are combined, they achieve optimal detection performance, leveraging feature separability and noise reduction to effectively counter both types of backdoor attacks.

\begin{table}[h]
\centering
\captionsetup{font=small}
\caption{Ablation study with different parts of SAM-enhanced PSD under various backdoor attacks and base poisoned sample detection on CIFAR-10 and ResNet18}
\label{tab:abla}
\renewcommand\arraystretch{1}
\scalebox{0.9}{
\begin{tabular}{l|cc|cc|cc}
\toprule
\multirow{2}{*}{Detection} & \multirow{2}{*}{SAM} & \multirow{2}{*}{FS} & \multicolumn2{c|}{SS} & \multicolumn2{c}{Beatrix} \\
  &  &  & TPR & FPR & TPR & FPR \\
\midrule
\multirow{4}{*}{BadNets} & \ding{53} & \ding{53} & 70.8 & 2.4 & 56.6 & 5.0 \\
  & \ding{51} & \ding{53} & 86.0 & 0.6 & 98.4 & 5.0 \\
  & \ding{53} & \ding{51} & 72.8 & 1.2 & 67.0 & 5.0 \\
  & \ding{51} & \ding{51} & 92.3 & 1.2 & 98.8 & 0.5 \\
\midrule
\multirow{4}{*}{Blended} & \ding{53} & \ding{53} & 32.9 & 4.4 & 5.0 & 5.0 \\
  & \ding{51} & \ding{53} & 90.6 & 0.3 & 79.8 & 5.0 \\
  & \ding{53} & \ding{51} & 60.4 & 1.9 & 27.1 & 5.0 \\
  & \ding{51} & \ding{51} & 94.6 & 1.1 & 99.8 & 1.5 \\
\bottomrule
\end{tabular}
}
\vspace{-0.2cm}
\end{table}

\begin{figure}[t]
    \centering
    \includegraphics[width=1.0\linewidth]{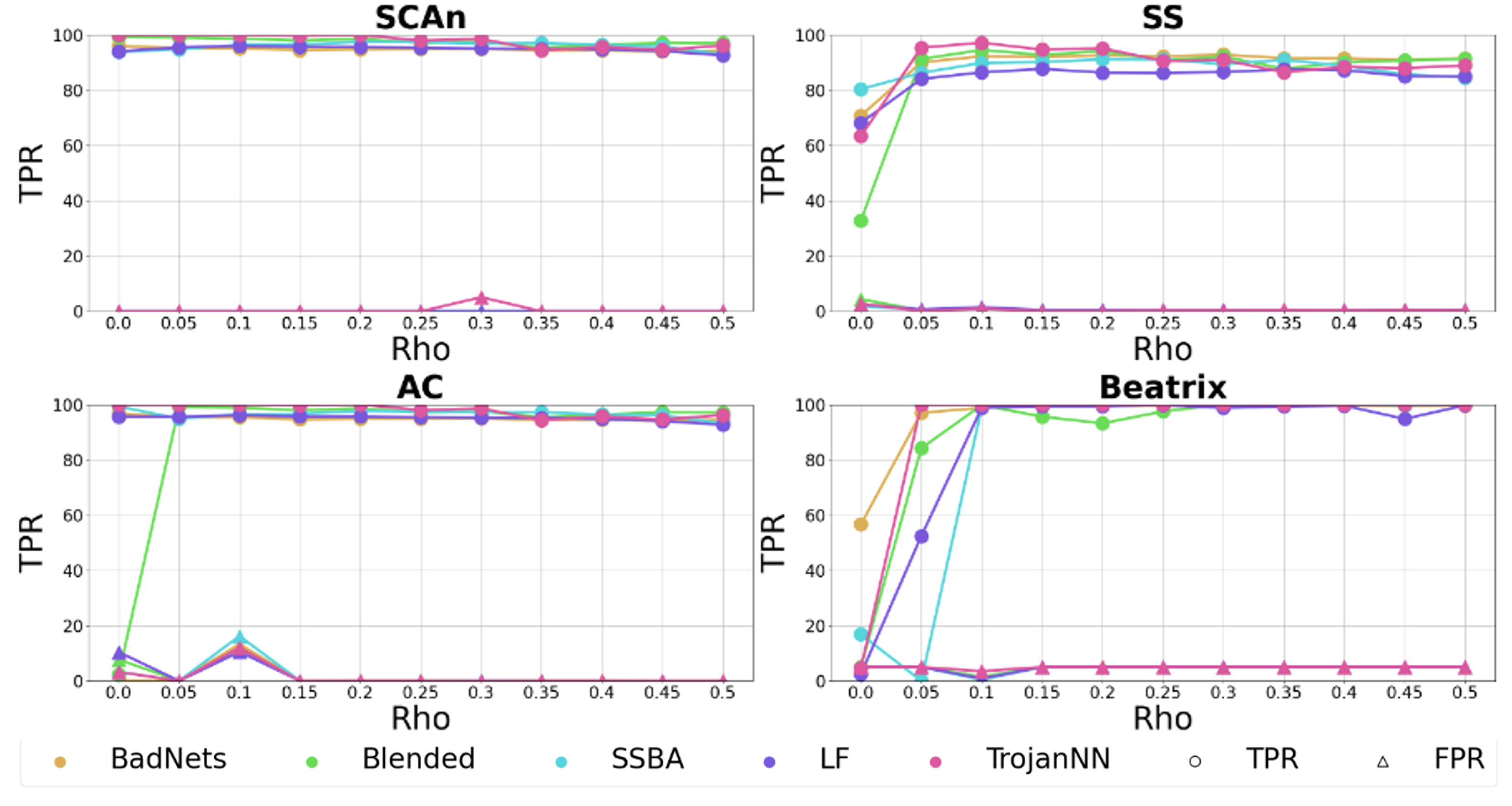}
    \vspace{-0.5cm}
    \caption{Detection performance of base PSD with SAM-enhanced PSD with different $\rho$ on CIFAR10 and ResNet18.}
    \label{fig:rho}
\vspace{-0.4cm}
\end{figure}

\paragraph{Detection performance with different $\rho$.}

The constraint bound $\rho$ is a key hyperparameter in our detection strategy, as it governs the extent of perturbation $\epsilon$. The excessively small $\rho$ may result in a weak enhancement of the backdoor effect, whereas an excessively large $\rho$ can degrade the model's performance by disrupting the extraction of information, such as features, from both poisoned and clean samples. We assess the sensitivity of $\rho$ by executing five complex attacks and employing four detection methods. Figure \ref{fig:rho} illustrates the results of these detections combined with SAM-enhanced PSD. While a smaller $\rho$ may not fully amplify backdoor effects resulting in poor performance of Beatrix against SSBA and TrojanNN attacks, it demonstrates that SAM-enhanced PSD can successfully identify poisoned samples and maintain a reasonable false positive rate across different $\rho$ settings. Overall, $\rho$ proves to be relatively insensitive; a broad range of values can be selected with good detection performance.

\subsection{Further Analysis}
\label{sec:ana}

\begin{figure}[t]
    \centering
    \includegraphics[width=1.0\linewidth]{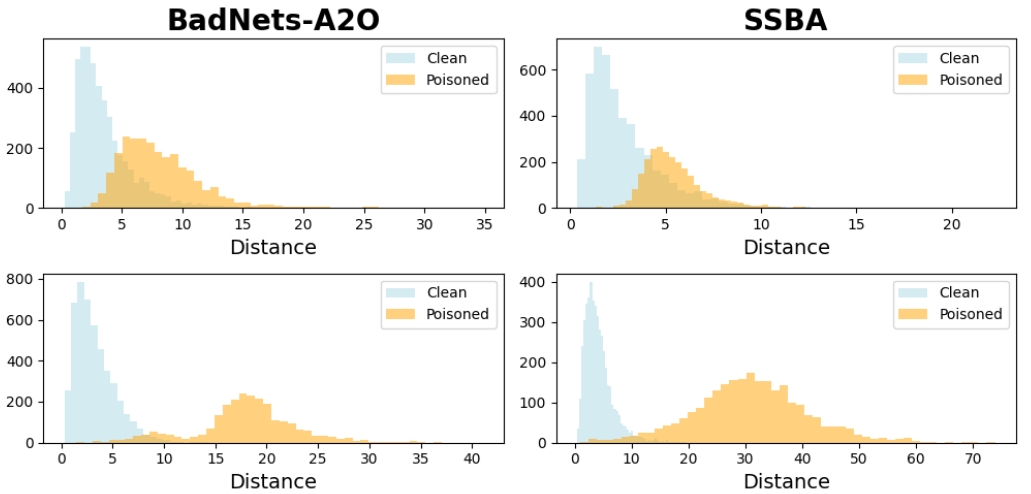}
    \vspace{-0.5cm}
    \caption{Distribution of distances between the target clean samples center and each samples in models trained with Vanilla training (Top row) and with SAM (Bottom row) under various backdoor attacks on CIFAR-10 and ResNet18.}
    \label{fig:feature}
\vspace{-0.5cm}
\end{figure}

\begin{figure*}[htbp]
    \centering
    \includegraphics[width=1.0\linewidth]{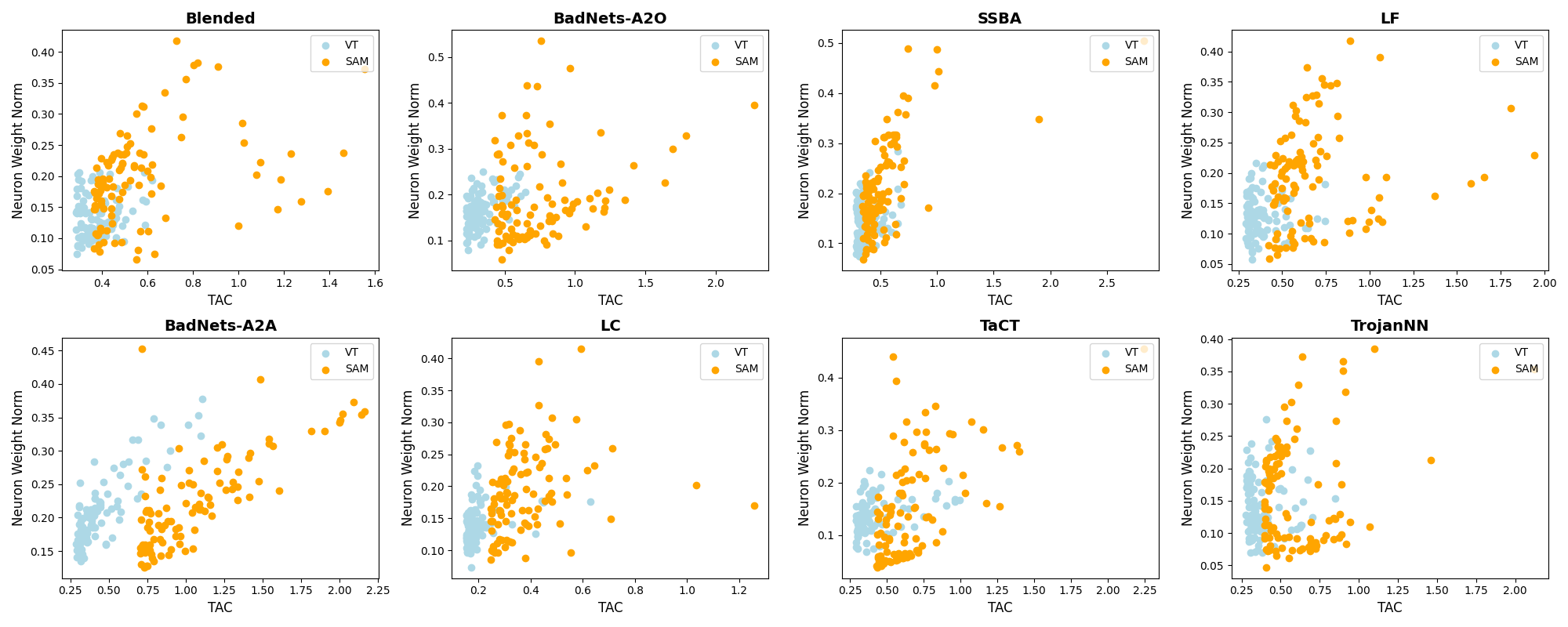}
    \vspace{-0.8cm}
    \caption{Neuron weight norm combined with TAC between the model trained with SAM and the model trained with Vanilla training under various backdoor attack on CIFAR-10 and ResNet18. The top 20\% of neurons ranked by TAC are displayed.}
    \label{fig:weight_norm}
\vspace{-0.4cm}
\end{figure*}

\begin{figure}[t]
    \centering
    \includegraphics[width=1.0\linewidth]{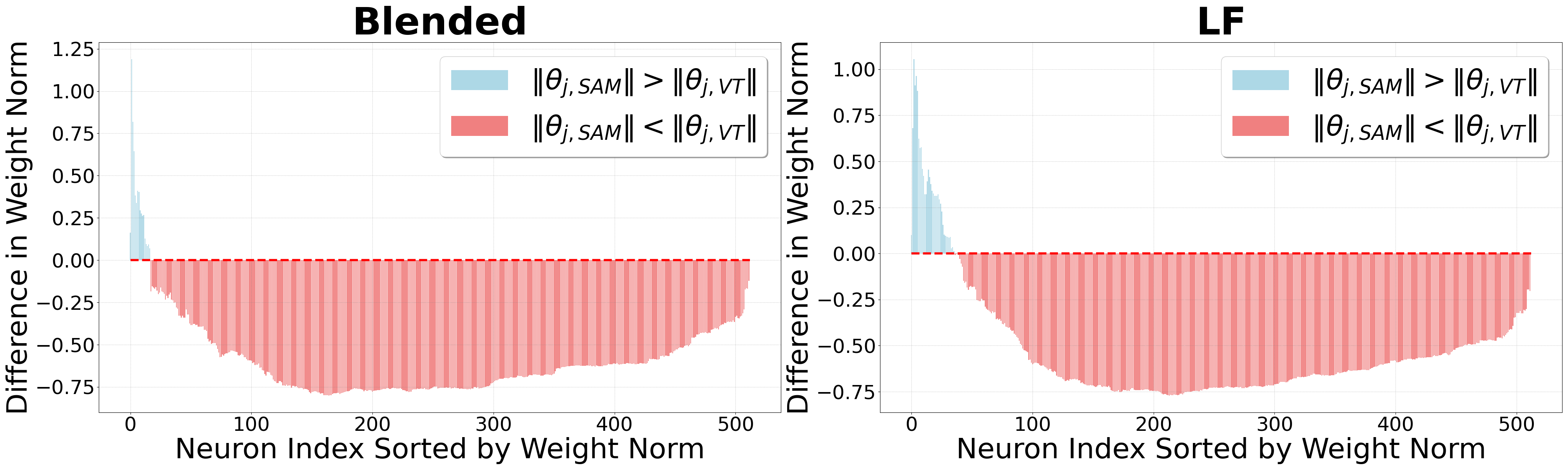}
    \vspace{-0.5cm}
    \caption{The differences in all weight norms between the model trained with SAM and the model trained with Vanilla Training. These neurons are indexed in descending order based on the weight norm in their respective models, which means that a smaller index indicates that this pair of neurons has a higher weight norm in the corresponding model.}
    \label{fig:weightdiff}
\vspace{-0.3cm}
\end{figure}

\paragraph{Visualization of samples feature space.}

To gain a deeper understanding of the impact of the SAM on poisoned and clean samples, we conduct a detailed analysis and visualization of the distribution of samples in the feature space across three dimensions. \ding{182} Distribution of distances between the target clean sample center and each sample: The center of the target clean samples is defined by the average of their features within the current model. As illustrated in \cref{fig:feature}, in models trained with Vanilla training (SGD), the features of poisoned samples are closer to the center of clean samples compared to those in models trained with SAM. Additionally, their distances to the center are smaller, which increases the difficulty in distinguishing between poisoned and clean samples. \ding{183} Silhouette Value Analysis: The silhouette coefficient is a metric that measures the cohesion within a class and separation between classes \citep{silhouette}. The silhouette coefficient for the BadNets attack improved from 0.19 in Vanilla training-trained models to 0.32 in SAM-trained models, and for the SSBA attack, it increased from 0.28 to 0.54 under the same conditions. This indicates that SAM-trained models are more effective in enhancing class separation, thereby aiding in better differentiation between classes. \ding{184} t-SNE Visualization: In the \textbf{supplementary material}, we demonstrate, through t-SNE, the distribution of poisoned and clean samples in the feature space, in which after training with the SAM algorithm, poisoned samples are more distinctly separated from the target class.

\paragraph{Weight Norms and TAC.}
To investigate the specific impact of SAM on the parameters of DNNs, we examined the distribution of neuron weight norms and their corresponding TAC. In \cref{fig:weightdiff}, we present the top 20\% of neurons with the highest TAC in the last convolutional layer of ResNet18 under different attack scenarios. As mentioned in FT-SAM \citep{ftsam}, there is a positive correlation between TAC and weight norm. When SAM is used for training instead of Vanilla Training, the weight norms of neurons with high TAC increase, and their corresponding TAC values also significantly rise. This suggests that SAM enhances the activation values of poisoned samples by increasing the weights of backdoor neurons. This is contrary to the phenomenon described in FT-SAM, where the weight norms of backdoor neurons decrease. This discrepancy arises because we train the model with poisoned samples, which is different from the approach used in FT-SAM. As shown in \cref{fig:weightdiff} (assuming the correct figure reference), the model predominantly increases the weight norms of neurons with larger norms, specifically the backdoor-related neurons, to effectively learn from the poisoned samples in the training set.

\section{Conclusion}

This work revisits existing poisoned sample detections (PSD) and finds that 
existing PSD methods often struggle with unstable performance when dealing with weak backdoor attacks, such as using low poisoning ratios or weak trigger strengths. Our statistical analysis reveals a positive correlation between the strength of the backdoor effect and the effectiveness of detection methods. Taking advantage of this insight, we propose to amplify the backdoor effect by training the model using Sharpness-Aware Minimization (SAM), without the need to change the poisoning ratio or trigger strength, thereby making poisoned samples more detectable. Our method, called \textit{SAM-enhanced PSD}, can be easily integrated with any existing PSD method that extracts discriminative features from the trained backdoor models for detection. Extensive experiments on various benchmark datasets and network architectures show that our method significantly improves detection performance. This work offers a valuable contribution to defending against backdoor attacks in deep neural networks, providing a new perspective that complements existing detection methods and has the potential to inspire further research in this critical area.

\newpage
{
    \small
    \bibliographystyle{ieeenat_fullname}
    \bibliography{main}
}


% \input{tables/cifar0.005}
% \input{tables/cifar0.01}
% \input{tables/cifar0.001}
% \input{sec/proof}

\end{document}